\documentclass{article}

\usepackage[final, nonatbib]{corl_2020} %

\usepackage{hhline}
\usepackage{amsmath}
\usepackage{amsfonts}
\usepackage{xcolor}
\usepackage{bm}
\usepackage{wrapfig}

\usepackage[natbib=true, style=numeric]{biblatex}
\usepackage{hyperref}       %
\usepackage{url}            %

\usepackage{enumitem, hyperref}
\makeatletter
\def\namedlabel#1#2{\begingroup
    #2%
    \def\@currentlabel{#2}%
    \phantomsection\label{#1}\endgroup
}
\makeatother

\usepackage[page,toc,titletoc,title]{appendix}

\usepackage{graphicx}
\usepackage{subcaption}
\usepackage{nicefrac}       %
\usepackage{microtype}      %
\usepackage{color}
\usepackage{float}

\usepackage[linesnumbered,ruled]{algorithm2e}
\usepackage{tabularx}
\usepackage{bbm}
\usepackage{amssymb}

\usepackage{bbm}
\usepackage{bm}
\usepackage{arydshln}

\usepackage{booktabs}

\usepackage[bottom]{footmisc}

\usepackage[subtle]{savetrees}

\usepackage[compact]{titlesec}
\titlespacing{\section}{0pt}{2ex}{1ex}
\titlespacing{\subsection}{0pt}{1ex}{0ex}
\titlespacing{\subsubsection}{0pt}{0.5ex}{0ex}

\usepackage[font=footnotesize,labelfont=bf]{caption}

\newcolumntype{R}{>{\hsize=.5\hsize\raggedleft\arraybackslash}X}
\newcolumntype{s}{>{\hsize=.5\hsize}X}

\newcommand{\pete}{\textcolor{blue}{Pete: }\textcolor{blue}}

\definecolor{MyRed}{rgb}{1.0,0.0,0.0}
\newcommand{\yl}[1]{\textcolor{MyRed}{[Yunzhu: #1]}}

\renewcommand{\yl}[1]{}
\renewcommand{\pete}[1]{}

\title{Keypoints into the Future: Self-Supervised Correspondence
in Model-Based Reinforcement Learning
}

\author{
  Lucas Manuelli\thanks{CSAIL, Massachusetts Institute of Technology}\\
  \texttt{manuelli@mit.edu} \\
  \And
  Yunzhu Li\footnotemark[1]\\
  \texttt{liyunzhu@mit.edu} \\
  \And
  Pete Florence \thanks{Robotics at Google}\\
  \texttt{peteflorence@google.com} \\
  \And
  Russ Tedrake\footnotemark[1]\\
  \texttt{russt@mit.edu} \\
}

\begin{document}
\maketitle

\begin{abstract}
Predictive models have been at the core of many robotic systems, from quadrotors to walking robots. However, it has been challenging to develop and apply such models to practical robotic manipulation due to high-dimensional sensory observations such as images. Previous approaches to learning models in the context of robotic manipulation have either learned whole image dynamics or used autoencoders to learn dynamics in a low-dimensional latent state. In this work, we introduce model-based prediction with self-supervised visual correspondence learning, and show that not only is this indeed possible, but demonstrate that these types of predictive models show compelling performance improvements over alternative methods for vision-based RL with autoencoder-type vision training. Through simulation experiments, we demonstrate that our models provide better generalization precision, particularly in 3D
scenes, scenes involving occlusion, and in category-generalization. Additionally, we validate that our method effectively transfers to the real world through hardware experiments. \href{https://sites.google.com/view/keypointsintothefuture}{https://sites.google.com/view/keypointsintothefuture}.
\end{abstract}


\section{Introduction}
\label{key_dynam:sec:intro}

It has been argued that one of the hallmarks of human-level learning is the ability to construct and leverage causal models of the world \cite{lake2017building}. In the area of manipulation, this manifests itself in the ability to approximately predict how an object will move if we grasp or push it. Traditional model-based robotics has successfully leveraged such predictive models, oftentimes derived from first principles, to solve challenging planning and control problems \citep{moore2014robust, mellinger2012trajectory}. In the area of practical vision-based robotic manipulation, however, it has been particularly hard to leverage such predictive models, due to varied and novel objects and the high-dimensional observation spaces involved (e.g., RGB or RGBD images). %
Alternative approaches, such as imitation learning or model-free reinforcement learning, sidestep the task of building a predictive model and directly learn a policy. Although this can be appealing, model-based techniques offer several benefits. They can be sample efficient compared to model-free methods and, in contrast to behavior cloning techniques, can leverage off-policy non-expert data. Once a model has been acquired, it can be used together with a planner to achieve a wide variety of tasks and goals. One of the main challenges for model learning applied to robotic manipulation is determining the state representation on which the dynamics model should be learned. Prior work has used approaches ranging from full image space dynamics \citep{finn2016unsupervised, ebert2017self, yen2019experience, suh2020surprising} to a variety of autoencoder formulations \citep{agrawal2016learning, hafner2018learning}. 

In this paper we propose to use object keypoints, which are tracked over time, as the latent state in which to learn the dynamics.  
These keypoints anchor our model-based predictions, and provide various advantages over alternatives such as abstract latent states: (i) the output is interpretable, which enables the ability to analyze the performance of the visual model separately from the predictive model. (ii) The representation is 3D and hence can naturally handle changing general off-axis 3D camera positions. (iii) As demonstrated in \citep{florence2018dense, florence2019self} the visual models we use, \textit{Dense Object Nets}, have demonstrated reliable performance in a variety of real-world settings and are able to generalize at the category level. We found that autoencoder approaches particularly struggled with category-level generalization. And as discussed in prior works \cite{florence2018dense,manuelli2019kpam}, keypoints and dense correspondences provide advantages over using 6D object poses: they can apply to deformable objects and represent category-level generalization.

We show that this formulation enables reliable, sample-efficient learning capable of precise visual-feedback-based manipulation in the real world -- and is trained with nothing other than a small amount of interaction data (10 minutes) and a single demonstration for goal specification.  In our approach to acquiring keypoints, we extract them as descriptors which are tracked from a dense descriptor model -- while multiple approaches could be used to acquire keypoints, this route can be entirely self-supervised.
As opposed to \citep{florence2019self} which uses keypoints from a dense-correspondence model in an imitation learning framework, the use of keypoints as input to a model-based RL system presents several unique challenges. In particular the keypoints need to be both informative for the task at hand and be able to be tracked accurately. In this paper we explore these challenges and propose solutions.

\begin{figure}[t]
\centering
\includegraphics[width=0.85\textwidth]{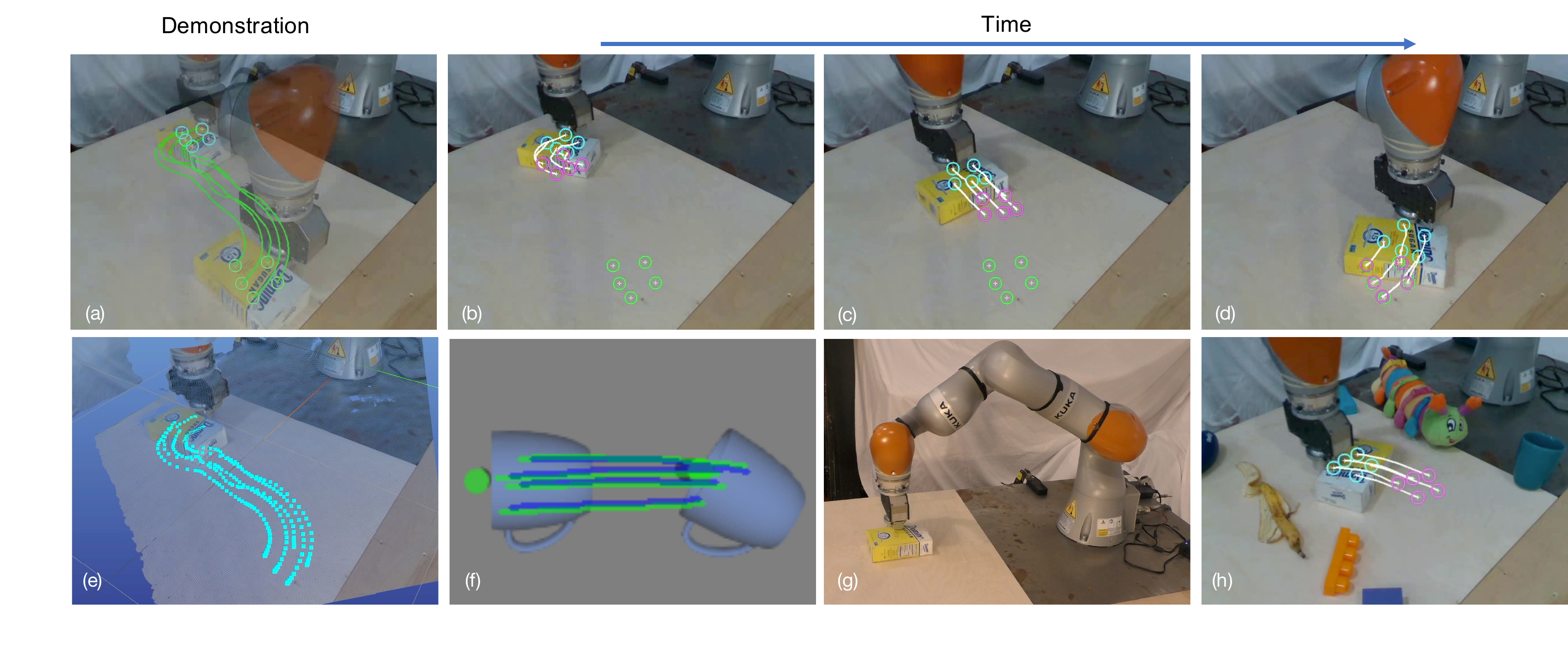}
\caption{\label{key_dynam:fig:figure_1} (a) Shows the initial pose (blue keypoints) and goal pose (green keypoints) along with the demonstration trajectory. (b) - (d) show the MPC at different points along the episode. Current keypoints are in blue, white lines and purple keypoints show the optimized trajectory from the MPC algorithm, using the learned dynamics model. Goal keypoints are still shown in green. (e) shows the demonstration trajectory in a 3D visualizer. (f) Illustrates a dynamics model on a category level task. The actual keypoint trajectory is shown in green; the predicted trajectory using the learned model shown in blue. (g) Overview of our hardware setup. (h) Example of visual clutter.}
\vspace{-0.6cm}
\end{figure}

\textbf{Contributions:}
Our primary contribution is (i) a novel formulation of predictive model-learning using learned dense visual descriptors as the state representation. We use this approach to perform closed-loop visual feedback control via model-predictive-control (MPC). (ii) Using simulated manipulation experiments we demonstrate that this approach offers performance benefits over a variety of baselines, and (iii) we validate our approach in real-world robot experiments. 

\section{Related Work}
We focus our related work on methods that target robotic manipulation with learned predictive models. The Introduction discussed alternative approaches to synthesizing closed-loop feedback controllers without predictive models, via imitation learning or model-free reinforcement learning. %

\textbf{Model-based RL in Robotic Manipulation.} These methods can be classified by whether they use first-principles-based \yl{will ``physics-based" be better?} or data-driven models, and whether they consume raw visual inputs (such as RGBD images) or they consume ground truth state information (from a simulator or an external perception system). %
First-principles-based models, e.g., \citep{hogan2016feedback, zhou2019pushing}, rely on known object models and thus don't generalize to novel or unknown objects, and also rely on external vision systems. %
Given this, the tasks we consider are out of scope for these approaches.  
In the area of methods that use external vision systems but learn data-driven models, \cite{hogan2018data} learns a dynamics model for a closed-loop-controlled planar pushing task, however, the approach is tailored to the specifics of the pusher-slider task and doesn't readily generalize to other tasks, or to novel objects within the same task.
\cite{nagabandi2019deep} learns deep dynamics models for a variety of different simulated dexterous manipulation tasks, using ground truth object states, and one task on real hardware, using an external camera-based 3D object position tracker. %
Although \citep{hogan2018data, nagabandi2019deep} achieve impressive results, their reliance on ground truth object state and/or specialized visual trackers limits their general applicability in more diversified real-world manipulation tasks.

There also exists a large literature on model-based RL for robotic manipulation that operates in the more challenging problem class of directly consuming image observations. %
Methods can be broadly categorized into whether they predict the image-space dynamics of the entire image \citep{finn2016unsupervised, ebert2018visual, ebert2017self, yen2019experience, suh2020surprising}, or predict the dynamics of a low-dimensional latent state  \citep{yan2020learning, watter2015embed, agrawal2016learning, hafner2018learning}. %
Although image-space dynamics approaches are general, they require large training datasets. %
For latent-space dynamics approaches,
to avoid a trivial solution where all observations get mapped to a constant vector, a regularization strategy is needed for the latent state $\bm{z}$. \citep{watter2015embed, hafner2018learning} use an autoencoder architecture and regularize the latent state $\bm{z}$ using a reconstruction loss. \citep{agrawal2016learning} regularize the latent space by simultaneously training both forward and inverse dynamics models while \cite{yan2020learning} uses a contrastive loss on the latent state. In contrast to these approaches we use visual-correspondence pretraining to produce a latent state which is physically grounded and interpretable as the 3D locations of keypoints on the object(s).

\textbf{Visual Representation Learning} \yl{will ``Visual Representation Learning for Closed-loop Control" be better?}%
For more related-work in self-supervised visual learning for robotics, we refer the reader to \cite{florence2018dense}. Approaches that use autoencoders \citep{watter2015embed, hafner2018learning} or full image-space dynamics \citep{finn2016unsupervised, ebert2018visual, ebert2017self, yen2019experience} rely on image reconstruction as their source of visual supervision. \cite{kulkarni2019unsupervised} is perhaps most related to ours, in that they first learn a visual model which is then used for a downstream task, and they show that freezing the visual model and using a keypoint-type representation as an input to model-free RL algorithms improves performance on Atari ALE \cite{bellemare2013arcade}. Our approach is distinct in that (i) we use a predictive model-based method rather than a model-free method, (ii) we use a correspondence-based training loss, while \cite{kulkarni2019unsupervised} uses an image-reconstruction-based loss with a specialized pixel-space transport mechanism, and (iii) we demonstrate results with real-world hardware.  As a baseline, we try using their vision model as an input to the same model-based RL algorithm used by our own method.

\section{Formulation: Self-Supervised Correspondence in Model-Based RL}
This section describes our approach to model-based RL using visual observations. The goal is to learn a dynamics model that can then be used to perform online planning for closed-loop control.

\subsection{Model-Based Reinforcement Learning}

Our setting consists of an environment with states $\bm{x} \in \mathcal{X}$, observations $\bm{o} \in \mathcal{O}$, actions $\bm{a} \in \mathcal{A}$ and transition dynamics $\bm{x}' = f_\text{state}(\bm{x}, \bm{a})$. The task is specified by a reward function $r(x,a)$ and the goal is to choose actions to maximize the expected reward over a trajectory. We approach this problem by first learning an approximate dynamics model $\hat{f}_{\theta_{\text{dyn}}}$, which is trained to minize the dynamics prediction error on the observed data $\mathcal{D}$. The learned model is then used to perform online planning to obtain a feedback controller.

Typical example environments are depicted in Figures \ref{key_dynam:fig:figure_1}, \ref{key_dynam:fig:sim_images}. The state $\bm{x}$ contains information about the underlying pose and physical properties of the object, but we only have access to the observation $\bm{o}$. The observation typically consists of both the robot's proprioceptive information $\mathcal{O}_\text{robot}$ (such as joint angles, end-effector poses, etc.) and high-dimensional images $\mathcal{O}_\text{image} \in \mathbb{R}^{W \times H \times C}$ for a $C$-channel image of height $H$ and width $W$. Hence the full observation space is $\mathcal{O} = \mathcal{O}_\text{robot} \times \mathcal{O}_\text{image}$. For the purposes of our approach we will assume that $\bm{x}$ is fully observable from $\bm{o}$ (or a short history of $\bm{o}$ in order to infer velocity information). Note that this still allows for the object to undergo significant partial occlusion as long as it is not completely occluded, see Figure \ref{key_dynam:fig:sim_images} (c) for an example. %
While our work focuses on addressing partial occlusion, future work may address full occlusion via models with longer time horizons or higher-level planning.

Rather than learn the dynamics directly in the observation space, as in \citep{finn2016unsupervised, ebert2018visual}, we instead learn a mapping $g:\mathcal{O} \to \mathcal{Z}$ from the high-dimensional observation space $\mathcal{O}$ to a low-dimensional latent space $\mathcal{Z}$ together with a dynamics model $\hat{\bm{z}}_{t+1} = \hat{f}_{\theta_{dyn}}(\bm{z}_{t-l:t}, \bm{a}_t)$ in this latent space, where $\bm{z}_{t-l:t} = (\bm{z}_{t}, \bm{z}_{t-1}, \ldots, \bm{z}_{t-l})$ encodes a short history (we use $l=1$ in all experiments). This latent state should capture sufficient information about the true state $\bm{x}$ such that driving $\bm{z} \to \bm{z*}$ sufficiently well achieves the goal of driving $\bm{x}$ to $\bm{x}^*$ (where $\bm{z}^*$ is the latent state corresponding to $\bm{x}^*$). Given the current latent state %
and a sequence of actions $\{\bm{a}_{t}, \bm{a}_{t+1}, \ldots\}$ we can predict future latent states $\bm{z}$ by repeatedly applying our learned dynamics model. The forward model is then trained to minimize the dynamics prediction error (also know as \textit{simulation error}) over a horizon $H$
\begin{equation}
\label{key_dynam:eq:dynamics_cost_function}
    \mathcal{L}_\text{dynamics} = \sum_{h=1}^H ||\hat{\bm{z}}_{t+h} - \bm{z}_{t+h}||_2^2, \hspace{0.1in} \hat{\bm{z}}_{t+h+1} = \hat{f}_{\theta_{\text{dyn}}}(\hat{\bm{z}}_{t-l:t}, \bm{a}_{t+h}), \hspace{0.1in} \hat{\bm{z}}_{t} = \bm{z}_t
\end{equation}
\subsection{Learning a Visual Representation}
\label{key_dynam:subsec:learning_visual_representation}

The objective of the visual model $g: \mathcal{O} \to \mathcal{Z}$ is to produce a low-dimensional feature vector $\bm{z}$ which serves as a suitable latent state in which to learn the dynamics. For the types of tasks and environments that we are interested in, spatial information about object locations is a critical piece of information. Pose estimation has played a critical role in classical manipulation pipelines, and was also used in dynamics learning approaches such as \citep{hogan2018data, nagabandi2019deep}. In general producing pose information from high-dimensional observations (such as RGBD images) requires a dedicated perception system. Although pose can be a powerful state representation when dealing with a single known object, as noted in \cite{florence2018dense, manuelli2019kpam, florence2019self} it has several drawbacks that limit its usefulness in more general manipulation scenarios. In particular it doesn't readily (i) extend to the case of deformable objects, (ii) generalize to novel objects or (iii) extend to category-level tasks.

Our strategy is to leverage visual-correspondence pre-training to track points on the object of interest. The locations of these tracked points can then serve as the latent state on which we learn the dynamics. Similar to the approach taken in \cite{florence2018dense, florence2019self}, we use visual-correspondence learning, which is trained in a completely self-supervised fashion, to train a visual model which that can be used to find correspondences across RGB images. We then propose several approaches to produce a low-dimensional latent $\bm{z}$ using the pre-trained dense-correspondence model.

First we give a bit of background on dense correspondence models (see \cite{ florence2019thesis, florence2018dense,} for more details). Given an image observation $\bm{o}_\text{image} \in \mathbb{R}^{W \times H \times C}$ (where $C$ denotes the number of channels), the dense-correspondence model $g_{\theta_\text{dc}}$ outputs a full-resolution descriptor image $\mathcal{I}_D \in \mathbb{R}^{W \times H \times D}$. Since we want to learn a dynamics model on a low-dimensional state, we need a way to construct $\bm{z}$ from the descriptor image $\mathcal{I}_D$. The idea, similar to \cite{florence2019self}, is for $\bm{z}$ to be a set of points on the object(s) that are localized in either image-space or 3D space. These points are represented as a set $\{d_i\}_{i=1}^K$ of K descriptors, where each $d_i \in \mathbb{R}^D$ is a vector in the underlying descriptor space. A parameterless \emph{correspondence function} $g_c(\mathcal{I}_D, d_i)$\footnote{see Appendix \ref{key_dynam:appendix:subsec:correspondence_function} for more details on the visual-correspondence model} extracts the location of the keypoint $y_i \in \mathbb{R}^B$ from the current observation. Combining our learned correspondences together with the reference descriptors, we have a function that maps image observations $\bm{o}_\text{image}$ to keypoint locations $\bm{y} = \{y_i\}_{i=1}^K$. We propose four methods for constructing the latent state $\bm{z} = g_{\theta_z}(\bm{y})$ from $\bm{y}$, where $\theta_z$ denotes the (potentially empty) set of trainable parameters in this mapping. 

\noindent
\textbf{Descriptor Set (DS):}
In our simplest variant the latent state $\bm{z}$ is simply made up of keypoint locations $y_i$ for a set of descriptor keypoints randomly sampled from the object. Specifically, we sample $K$ (we use $K = 50$ in all experiments) descriptors $\{d_i\}_{i=1}^K$ corresponding to pixels from a masked reference descriptor image in our training set. Thus
\begin{align}
    \bm{z} 
    &= (\bm{z}_\text{object}, \bm{o}_\text{robot}) = (\bm{y}, \bm{o}_\text{robot}) = (\{y_i\}_{i=1}^K, \bm{o}_\text{robot})
\end{align}

\begin{wrapfigure}{R}{0.6\textwidth}
\vspace{-1.6em}
  \begin{center}
    \includegraphics[width=0.58\textwidth]{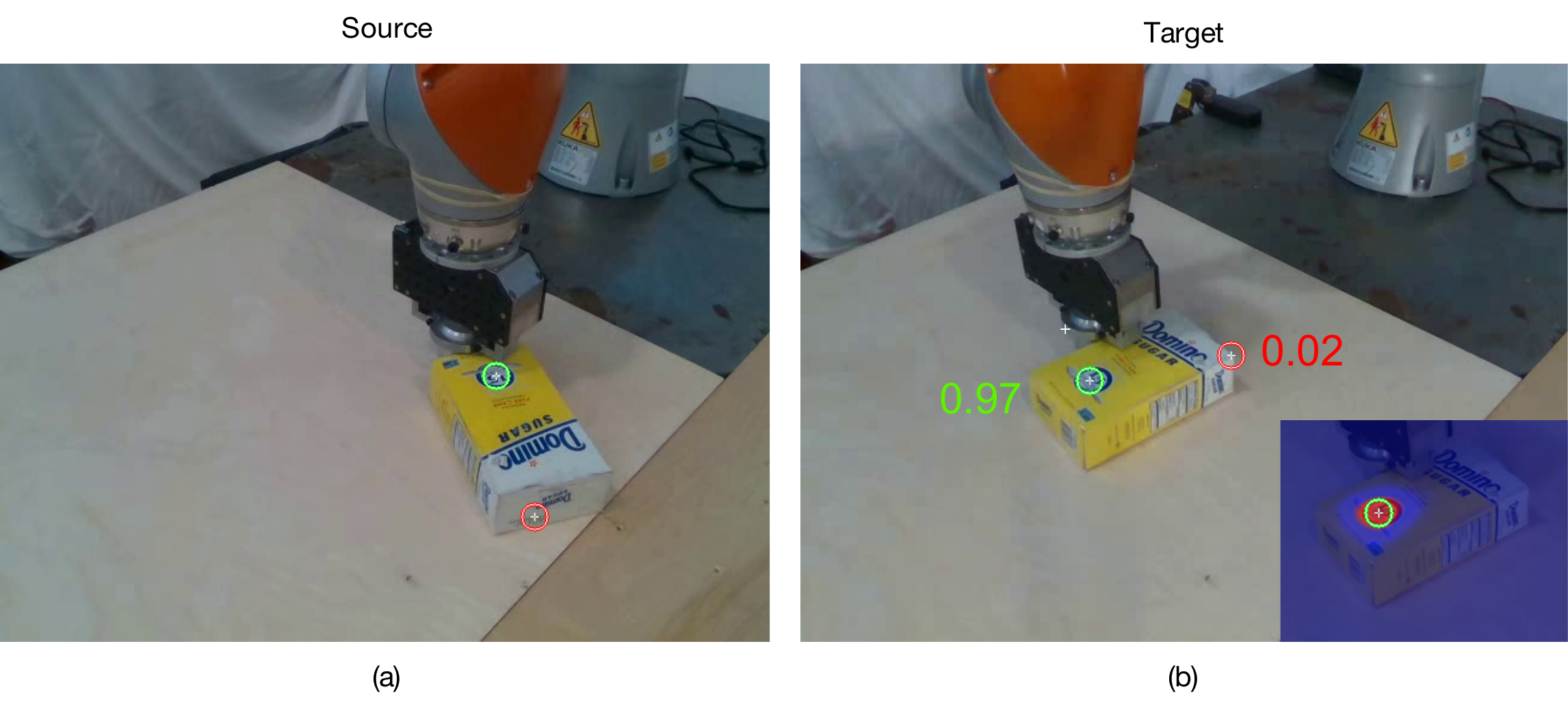}.
  \end{center}
  \caption{\label{key_dynam:fig:descriptor_confidence} 
 Visualization of the learned visual-correspondence model on a reference image (left) and target image (right). %
 Colored numbers in the target image represent the probability that the detected correspondence is valid. Green reticle shows a valid correspondence with a high confidence score, red reticle shows a case where no correspondence exists in the target image due to occlusion, hence the low confidence probability. Confidence heatmap shown in bottom right.}
 \vspace{-0.5cm}
\end{wrapfigure}

\noindent
\textbf{Spatial Descriptor Set (SDS):}
Rather than randomly sampling descriptors, as in \textbf{(DS)}, this method attempts to choose descriptors $\bm{d} = \{d_i\}_{i=1}^K$ having specific properties. In particular we would like the descriptors $\{d_i\}_{i=1}^K$  to be (i) \emph{reliable}, and (ii) \emph{spatially separated}. By \emph{reliable} we mean that they can be localized with high accuracy and don't become occluded during the typical operating conditions, see Figure \ref{key_dynam:fig:descriptor_confidence} for an example. \emph{Spatially separated} means that the chosen descriptors aren't all clustered around the same physical location on the object(s) of interest, but rather are sufficiently spread out (either in 3D space or pixel space) to provide meaningful information about both object position and orientation. Our dense descriptor model can provide a confidence score associated with descriptors and their associated correspondences.\footnote{see Appendix \ref{key_dynam:appendix:subsec:correspondence_function} for more details} Figure \ref{key_dynam:fig:descriptor_confidence} shows a clear example of high confidence for a valid match and low-confidence when no valid correspondence exists due to an occlusion. We use this feature of our visual model to compute a confidence score $c_i$ for each each descriptor $d_i$ according to what fraction of images in the training set $\mathcal{D}$ contain a high probability correspondence for $d_i$. The intuition is that descriptors $d_i$ corresponding to points on the object that are easy to localize and remain unoccluded will have a high confidence score. As in the \textbf{DS} method we initially select a large number of descriptors ($K = 100$) corresponding to points on the object. 
We then select the $K^*$ descriptors with the highest average confidence on the training set and which additionally satisfy a threshold on minimum separation distance (typically $25$ pixels in a $640 \times 480$ image). In the experiments we use $K^* = 4$ or $K^*=5$.

\noindent
\textbf{Weighted Descriptor Set (WDS):}
One problem that can arise when learning the dynamics of a latent state $\bm{z}$ is that some component of $\bm{z}$ may be noisy or unreliable, which can make it difficult or impossible to learn a dynamics model $\bm{z}' = f(\bm{z}, \bm{a})$. This problem doesn't arise in the imitation learning setting of \cite{florence2019self} which performs supervised learning from $\bm{z}_t \to \bm{a}_t^\text{expert}$, and thus can learn to ignore components of the latent state $\bm{z}_t$ that aren't useful for predicting the action $\bm{a}_t^\text{expert}$. The fundamental difference of our dynamics learning formulation compared to the imitation learning setup of \cite{florence2019self} is that the latent $\bm{z}_t$ appears directly in the cost function \eqref{key_dynam:eq:dynamics_cost_function}. There are a variety of reasons that one or more of the tracked descriptors could be unreliable (e.g. occlusions, regions of the object where the correspondence model has less accurate) and thus we would like our dynamics learning framework to be robust to this. We achieve this by defining a learnable mapping $\phi: \bm{y} \to \bm{z}$ which maps the keypoint locations $\bm{y}$ to the latent state $\bm{z}$, where the keypoints $bm{y}$ are as in our \textbf{DS} method. A regularization strategy is needed to ensure that $\phi$ doesn't collapse to the trivial solution $\phi \equiv 0$.
We introduce learnable weights $\alpha \in \mathbb{R}^{K \times K}$ and define $ w_{k, i} = \frac{\exp(\alpha_{k,i})}{\sum_{j=1}^K \exp(\alpha_{k, j})}$. Let $W \in \mathbb{R}^{K \times K}$ be the matrix with entries $w_{k, i}$, where the parameterization guarantees that $w_{k, i} \geq 0$ and $\sum_{i=1}^K w_{k, i} = 1$. $\phi$ is defined as
\begin{align}
    \tilde{y}_k &= \sum_{i=1}^K w_{k, i} y_i, \hspace{0.1in} \phi(\bm{y}; \alpha) = \tilde{\bm{y}} = \{\tilde{y}_k\}_{k=1}^K
\end{align}
Note that each $y_i$ is a keypoint location in $\mathbb{R}^B$. Thus $\tilde{\bm{y}}$ is simply a convex combination of the keypoints in $\bm{y}$. The latent state $\bm{z}$ is then defined as
\begin{align}
    \bm{z} =(\bm{z}_\text{object}, \bm{o}_\text{robot})
    = (\tilde{\bm{y}}, \bm{o}_\text{robot})
    = (\phi(\bm{y}, \alpha), \bm{o}_\text{robot})
\end{align}
The learnable weights $\alpha$ are trained jointly with the parameters $\theta_\text{dyn}$ of the dynamics model, and are fixed at test time. The fact that $\bm{z}_\text{object}$ is gotten from $\bm{y}$ by taking a weighted linear combination preserves the interpretation of $\bm{z}$ as tracking keypoints on the object, while allowing some flexibility to ignore unreliable keypoints.

\noindent
\textbf{Weighted Spatial Descriptor Set (WSDS):}
This method is simply the combination of \textbf{(SDS)} and \textbf{(WDS)}.
\subsection{Learning the Dynamics}
We adopt a standard dynamics learning framework where we aim to predict the evolution of the latent state $\bm{z}$. We train our dynamics model to minimize the prediction error in Equation \eqref{key_dynam:eq:dynamics_cost_function}
where $\bm{z}_t = g_{\theta_z}(\bm{o}_t)$. Our proposed methods differ in the structure of the mapping $g: \mathcal{O} \to \mathcal{Z}$ and the set of trainable parameters $\Theta$.\footnote{see Appendix \ref{key_dynam:appendix:subsec:training_details} for details} For all of our methods we keep the weights of the visual-correspondence network, $\theta_{dc}$, fixed.

\subsection{Online Planning for Closed-Loop Control}
\label{key_dynam:subsec:online_planning}
Once we have learned a dynamics model $\bm{z}' = f(\bm{z}, \bm{a})$, we use online planning with MPC to select an action. Given a goal latent state $\bm{z}^*$, we want to find an action sequence $\{\bm{a}_{t'}\}_{t'=t}^{t+H-1}$ that maximizes the reward $R = \sum_{t'=t}^{t+H-1} r(\bm{z}_{t'}, \bm{a}_{t'})$. Our dynamics learning approach is agnostic to the type of optimizer used to solve the MPC problem and the focus of our work is on the visual and dynamics learning, rather than the specifics of the MPC. Many model-based RL approaches \citep{yen2019experience, ebert2017self, nagabandi2019deep, hafner2018learning, finn2016unsupervised} use a random-sampling based planner (e.g. cross-entropy method) to solve the underlying MPC problem. We experimented with random shooting, gradient based shooting, cross-entropy and model-predictive path integral (MPPI) planners and found that MPPI worked best in our scenarios.\footnote{See Appendix \ref{key_dynam:appendix:sec:mpc} for more details.}

\section{Results}

We perform experiments aimed at answering the following questions: (1) Is it possible to successfully use self-supervised descriptors as the latent state for a model-based RL system? (2) What is the effect of various design decisions in our algorithm? (3) How does visual-correspondence learning compare to several benchmark methods in terms of enabling effective model-based RL policies? (4) Can we apply the method on real hardware?  

Our main contribution is on the formulation of the visual model and dynamics learning problem rather than the specifics of the MPC. However, to accurately compare our approach to various baselines, we need to perform experiments in which the dynamics model is used in closed-loop to solve a manipulation task. Our formulation of dynamics learning is very general, and in principle can handle a wide variety of manipulation scenarios. However, even with an accurate model (whether it comes from first principles or is learned), using this model to perform closed-loop feedback control remains a challenging problem. Thus, for our closed-loop experiments, we limit ourselves to pushing tasks that can adequately be solved by the planners outlined in Section \ref{key_dynam:subsec:online_planning}.

\noindent
\textbf{Tasks:} Extended experimental details are provided in the Appendix but we provide a brief overview of the tasks here. We perform experiments with four simulated tasks (depicted in Figure \ref{key_dynam:fig:sim_images}) and one hardware task. All tasks involve pushing an object to a desired goal state. The first three simulation tasks, denoted as \textit{top-down, angled, occlusions} involve pushing a single object. \textit{top-down} has cameras in a top-down orientation while they are angled at 45 degrees in \textit{angled}. Task \textit{occlusions} keeps the angled camera positions but the box is resting on a different face, resulting in significant self-occlusions. Task \textit{mugs} involves pushing many different objects from a category, in this case mugs with different size, shape and textures. The hardware task is essentially identical to the \textit{angled} sim task.

\subsection{Visual-correspondence Performance}
\label{key_dynam:subsec:visual_correspondence_performance}

\begin{figure}[t]
\centering
\includegraphics[width=1.0\textwidth]{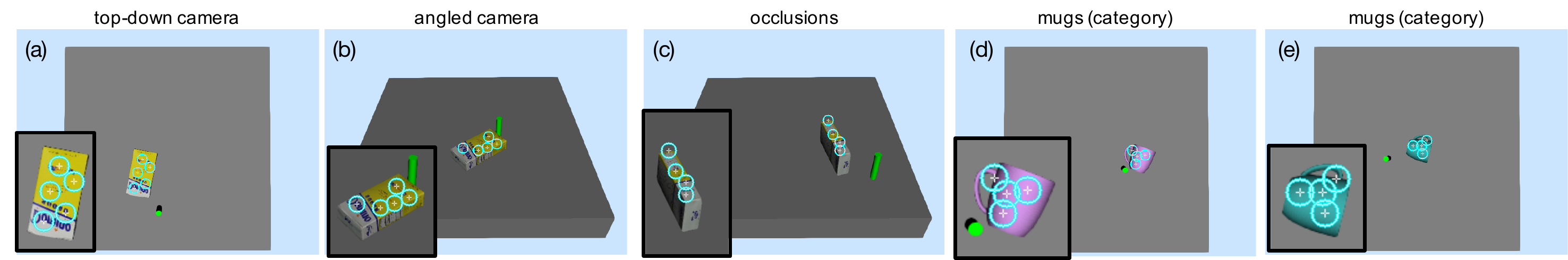}
\caption{\label{key_dynam:fig:sim_images} 
This figure shows reference descriptors $\{d_i\}_{i=1}^K$ of the \textbf{Spatial Descriptor Set} variant for our different simulation environments. In particular in (c) the sides of the object undergo occlusions as the box rotates about the vertical axis. (d)-(e) show two different mugs from the category-level \textit{mugs (category)} task.}
\vspace{-0.5cm}
\end{figure}

Figures \ref{key_dynam:fig:figure_1}, \ref{key_dynam:fig:descriptor_confidence} show the performance of our dense visual-correspondence model. In particular Figure \ref{key_dynam:fig:figure_1} shows the localization performance on real data along a trajectory while Figure \ref{key_dynam:fig:descriptor_confidence} shows an example of the confidence scores used as in the \textbf{SDS} method. Figure \ref{key_dynam:fig:sim_images} shows the descriptors used in the \textbf{SDS} method for each of our simulation tasks. In particular Figure \ref{key_dynam:fig:sim_images} (d)-(e) shows the ability of the descriptors to accurately find correspondences across different object instances within a category, despite differences in color and shape.

\subsection{Ablations on visual-correspondence for dynamics learning}
\label{key_dynam:subsec:ablations}

\begin{table}
\centering
\scriptsize{
\begin{tabularx}{\textwidth}{XRRRRRRRR}
\toprule
\textit{Task} & \multicolumn{2}{c}{\textit{top-down camera}} & \multicolumn{2}{c}{\textit{angled camera}} & \multicolumn{2}{c}{\textit{occlusions}} & \multicolumn{2}{c}{\textit{mugs (category)}} \\
  \cmidrule(lr){2-3} \cmidrule(lr){4-5} \cmidrule(lr){6-7} \cmidrule(lr){8-9}
         Method / task data &  pos, cm &  angle, $^\circ$ & pos, cm &  angle, $^\circ$ & pos, cm &  angle, $^\circ$ &  pos, cm &  angle, $^\circ$ \\
 \midrule
 Avg. trajectory  &  11.32 &  32.05 &  11.32 &  32.05 &  10.36 &  31.81 &  11.28 &  56.25  \\
 \midrule
 GT 3D &  1.14 &  4.01 &  1.14 &  4.01 &  1.29 &  5.45 & -- & -- \\
 SDS &  1.19 &  3.88 &  1.10 &  3.97 &  1.46 &  6.77 &  1.20 &  15.03 \\
 WDS &  1.16 &  4.43 &  1.30 &  5.12 &  1.49 &  8.64 &  1.00 &  13.29 \\
 DS &  1.12 &  4.07 &  1.45 &  5.50 &  3.66 &  12.27 &  1.19 &  11.73 \\
 WSDS &  1.19 &  4.00 &  1.18 &  4.86 &  1.28 &  5.59 &  1.39 &  18.18 \\
\bottomrule
\end{tabularx}
}
\caption{\textit{Ablations and comparison with ground-truth} -- quantitative results for various ablations of our method on four simulated tasks. Each method was evaluated on the same set of 200 different initial and goal states. The \emph{pos} and \emph{angle} columns denote the translational (in cm) and rotational (in degrees) deviations of the object from the goal position, averaged across all trials for a specific method and task. \emph{Avg. trajectory} denotes the average translation and rotation between the initial and goal poses for each task.\yl{bold the best performing numbers? also, the order: SDS, then WDS, then DS, then WSDS is a bit confusing.} \yl{listing ``goal states" as a ``method" also seems a bit weird to me}}
\label{key_dynam:table:sim_ablation_quant_results}
\vspace{-0.4cm}
\end{table}

Ablation studies show that the choice of \emph{what to track} can have a substantial impact on model performance, especially in the case of partial occlusions. 
Quantitative results are detailed in Table \ref{key_dynam:table:sim_ablation_quant_results}. On tasks \textit{top-down} and \textit{angled camera}  all methods perform reasonably well, almost matching the performance of the \textbf{GT 3D} baseline that uses ground truth state information. Intuitively this is because there are minimal occlusions in these settings, and so almost all keypoints can be tracked reliably using dense-visual-correspondence. In contrast the \textit{occlusions} introduces the potential for significant occlusions. Given the camera angle as shown in Figure \ref{key_dynam:fig:sim_images} (c) and the fact that the object rotates through the full $360$ degrees of yaw during the task, only the top face of the box remains unoccluded while the 4 side faces are alternately occluded and visible. This task exposes significant differences in performance among our various ablations. In particular \textbf{SDS}, \textbf{WDS} and \textbf{WSDS} perform significantly better than \textbf{DS}. We believe that this is due to the fact that some of the descriptors $\{d_i\}_{i=1}^K$ that are tracked in \textbf{DS} correspond to locations on the object that become occluded during an episode. The dense visual-correspondence model is not able to track keypoints through occlusions, and when trying to localize an occluded point our dense-correspondence model maps it to the closest point in descriptor space, which is not the location of the true correpondence. Hence the keypoint locations that makeup the latent state $\bm{z}_\text{object}$ for \textbf{DS} suffer reduced accuracy, leading to a less accurate dynamics model and ultimately lower performance when used for closed-loop MPC.

On the category-level task \textit{mugs} the camera is in a top-down position and thus occlusions are no longer an issue. However because the task involves different objects from a category there is shape variation among the different objects. Thus the methods that use $K=50$ keypoints, namely \textbf{DS} and \textbf{WDS}, perform better than the sparser variants \textbf{SDS}, \textbf{WSDS} that use only $K=4,5$ keypoints. We believe that this is due to the fact that having a larger number of keypoints better captures the shape variation across object instances and allows for a more accurate dynamics model.

\subsection{Comparison of visual-correspondence pretraining with baselines}

\begin{table}
\centering
\scriptsize{
\begin{tabularx}{\textwidth}{XRRRRRRRR}
\toprule
\textit{Task} & \multicolumn{2}{c}{\textit{top-down camera}} & \multicolumn{2}{c}{\textit{angled camera}} & \multicolumn{2}{c}{\textit{occlusions}} & \multicolumn{2}{c}{\textit{mugs (category)}} \\
  \cmidrule(lr){2-3} \cmidrule(lr){4-5} \cmidrule(lr){6-7} \cmidrule(lr){8-9}
         Method &  pos, cm &  angle, $^\circ$ & pos, cm &  angle, $^\circ$ & pos, cm &  angle, $^\circ$ &  pos, cm &  angle, $^\circ$ \\
\midrule
 SDS (ours) &  1.19 &  \textbf{3.88} &  \textbf{1.10} &  \textbf{3.97} &  \textbf{1.46} &  \textbf{6.77} &  1.20 &  15.03 \\
 WDS (ours) &  \textbf{1.16} &  4.43 &  1.30 &  5.12 &  1.49 &  8.64 &  \textbf{1.00} &  \textbf{13.29} \\
 Transporter 3D &  2.01 &  15.95 &  4.36 &  25.14 &  3.72 &  20.65 &  2.81 &  61.22 \\
 Transporter 2D &  2.08 &  13.59 &  3.60 &  23.60 &  2.74 &  18.86 &  2.33 &  60.18 \\
 Autoencoder &  2.18 &  14.79 &  2.85 &  13.79 &  3.08 &  14.20 &  9.05 &  56.84 \\
\bottomrule
\end{tabularx}
}
\caption{\textit{Comparisons with baselines} -- quantitative results of our method compared to various baselines on our four simulated tasks. Each method was evaluated on the same set of 200 different initial and goal states. \emph{pos} and \emph{angle} denote the translational (in cm) and rotational (in degrees) deviations of the object from the goal position, averaged across all trials for a specific method and task. %
}
\label{key_dynam:table:sim_baseline_quant_results}
\vspace{-0.5cm}
\end{table}
In our comparisons against baselines, our model outperforms alternatives on all experimental tasks.  The largest differences are apparent in tasks \textit{occlusions} and \textit{mugs} which involve partial occlusions and category-level generalization, respectively. The \textbf{transporter} baseline uses the keypoint locations from \cite{Kulkarni2019} as the latent state $z$ , while the \textbf{autoencoder} baseline jointly trains an autoencoder with a forward dynamics model. For a detailed discussion of the baselines see Appendix \ref{key_dynam:appendix:subsec:baselines}. Quantitative results are detailed in Table \ref{key_dynam:table:sim_baseline_quant_results}.

On task \textit{top-down camera}, the \textbf{transporter}\cite{kulkarni2019unsupervised} model was able to achieve performance that was only slightly worse than \textbf{WDS} and \textbf{SDS}, while the \textbf{autoencoder} performed significantly worse. The top-down setting is ideally suited for the feature transport approach of the transporter model. 

On tasks \textit{angled camera} and \textit{occlusions}, which have angled camera positions as opposed to the top-down task, the performance of our methods remained consistent while \textbf{transporter} suffered. This is potentially due to the fact that the feature transport mechanism of \textbf{transporter} is not well-suited to off-axis camera positions in 3D worlds. The performance of the \textbf{autoencoder} baseline remainded consistent, but worse than our approach, across tasks without category-level generalization.

Task \textit{mugs} contains a variety of different mug shapes with varied visual appearances and hence tests category-level generalization of both the perception and dynamics models. As discussed in Section \ref{key_dynam:subsec:visual_correspondence_performance}, our dense-correspondence model is able to find correspondences across these variations in appearance using only self-supervision, which allows us to learn a dynamics model that is effective for completing the task. This task is significantly harder than the other tasks not only because of the presence of novel objects, but because the goal states involve much larger rotations, as detailed in the first row of Table \ref{key_dynam:table:sim_ablation_quant_results}. Both the \textbf{transporter} and \textbf{autoencoder} baselines perform poorly in this task. We hypothesize that this is due to the fact that there is much more variance in the visual appearance of the objects as compared to the other tasks and thus the latent state $\bm{z}$ produced by these baselines is not amenable to dynamics learning.

\subsection{Hardware}
\label{key_dynam:subsec:hardware}
\noindent
\textbf{Experimental Setup:}
We used a Kuka IIWA LBR robot with a custom cylindrical pusher attached to the end-effector to perform our hardware experiments, see Figure \ref{key_dynam:fig:figure_1}. RGBD sensing was provided by two RealSense D415 cameras rigidly mounted offboard the robot and calibrated to the robot's coordinate frame. To enable effective correspondence learning between views, it is ideal to have views with \textit{some} overlap such that correspondences exist, but still maintain different-enough views. At test time only a single camera is used to localize the dense-descriptor keypoints. The robot is controlled by commanding end-effector velocity in the $xy$ plane at 5Hz.

\begin{figure}
\centering
\includegraphics[width=1.0\textwidth]{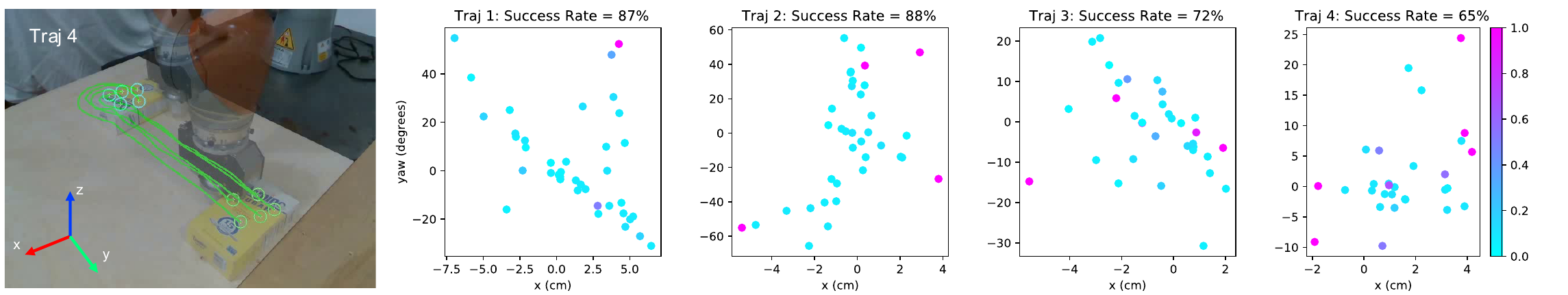}
\caption{\label{key_dynam:fig:hardware_results}
Left image shows demonstration for trajectory 4. Scatter plots show results of our approach on the four different reference trajectory tracking tasks. The axes of the plots show the deviation of the object starting pose from the initial pose of the demonstration. Color indicates the distance between final and goal poses, lower cost is better. The various reference trajectories are of different difficulties, as reflected by the different regions of attraction of the MPC controller.
More details can be found in Appendix \ref{key_dynam:appendix:hardware} and videos are on our \href{https://sites.google.com/view/keypointsintothefuture}{project page}}.
\vspace{-0.9cm}
\end{figure}

\noindent
\textbf{Hardware Results:}
For visual learning we collected a small dataset of the object in different positions to provide a diverse set of views for training the dense-correspondence model. For dynamics learning, we collected a dataset of the robot randomly pushing the object around. This amounted to approximately 10 minutes of interaction time and was used to train the dynamics model. All hardware experiments used the \textbf{SDS} method. To enable our MPC controller to accomplish long-horizon tasks, we supplied the controller with a reference trajectory for the object keypoints that came from a single demonstration, see Figure \ref{key_dynam:fig:figure_1} (a),(d). The MPC controller then tracked this reference trajectory using a $2$ second MPC horizon, which corresponds to $H=10$ since we are commanding actions at $5$ Hz. We collected 4 different reference trajectories\footnote{see Appendix \ref{key_dynam:appendix:hardware} for details on the reference trajectories} and ran multiple rollouts for each trajectory, varying the initial condition of the object pose during each run to test the region of attraction of our controller. In all cases our controller showed the ability to stabilize the system to the reference trajectory in spite of perturbations to the initial condition. Quantitative results are detailed in Figure \ref{key_dynam:fig:hardware_results}. In particular, we see that the ability of the controller to stabilize the trajectory in the face of disturbances in the initial condition depends on the trajectory. For trajectory (1) the controller is able to stabilize disturbances of up to 60 degrees, while trajectory (4) has a much smaller region of attraction. As can be seen in Appendix \ref{key_dynam:appendix:hardware}, trajectory (1) is a relatively simple trajectory with minimal orientation change between start and goal, while trajectory (4) involves a challenging 180-degree orientation change and requires the robot to operate at the edge of its kinematic workspace, reducing its control authority. Overall, our system exhibits impressive feedback and is able to track a trajectory in the keypoint latent space, enabling one-shot imitation learning. We encourage the reader to watch the videos on our \href{https://sites.google.com/view/keypointsintothefuture}{project page} to see the system in action.

\section{Conclusion}

We presented a method for using self-supervised visual-correspondence learning as input to a predictive dynamics model. Our approach produces interpretable latent states that outperform competing baselines on a variety of simulated manipulation tasks. Additionally, we demonstrated how the category-level generalization of our visual-correspondence model enables learning of a category-level dynamics model, resulting in large performance gains over baselines. Finally, we demonstrated our approach on a real hardware system, and showed its ability to stabilize complex long-horizon plans by tracking the latent state trajectory from a single demonstration.

\clearpage
\acknowledgments{This work was supported by Amazon.com Services LLC (Award No. CC MISC 00272683 2020 TR) and an Amazon Research Award. The views expressed are not endorsed by our funding sponsors.}

\printbibliography

\newpage

\label{key_dynam:appendix}

\section*{Appendix}
\section{Dense Correspondence}
\label{key_dynam:appendix:dense_correspondence}

Here we give a brief overview of the dense correspondence model formulation \cite{schmidt2017,florence2018dense} with spatial distribution losses \cite{florence2019self, florence2019thesis}. We briefly explain the loss functions and how the descriptors $\{d_i\}$ are localized.  

\subsection{Network Architecture}
We use an architecture that produces a full resolution descriptor image. Namely it maps
\begin{equation}
    W \times H \times 3 \rightarrow W \times H \times D
\end{equation}
We use a FCN (fully-convolutional network architecture) \cite{long2015fully} with a ResNet-50 or ResNet-101 with the number of classes set to the descriptor dimension.  Note that the FCN used in this work does not use striding and upsampling as in the architecture originally used in \cite{florence2018dense}.

\subsection{Loss Function}

For all shown experiments we use the spatially-distributed loss formulation with a combination of heatmap and 3D spatial expectation losses, as described in \cite{florence2019self}, Chapter 4.

\subsubsection{Heatmap Loss}

Let $p^*$ be the pixel space location of a ground truth match. Then we can define the ground-truth heatmap as
\begin{equation}
    H_{p^*}^*(p) = \exp\left(-\frac{||p - p^*||_2^2}{\sigma^2} \right)
\end{equation}
$p$ represents a pixel location. A predicted heatmap can be obtained from the descriptor image $\mathcal{I}_D$ together with a reference descriptor $d^*$. Then the predicted heatmap is gotten by
\begin{equation}
\label{key_dynam:appendix:eq:heatmap}
    \hat{H}(p; d^*, \mathcal{I}_D, \eta) = \exp\left(- \frac{||\mathcal{I}_D(p) - d^*||_2^2}{\eta^2} \right)
\end{equation}
The heatmap can also be normalized to sum to one, in which case it represents a probability distribution over the image.
\begin{equation}
    \tilde{H}(p) = \frac{\hat{H}(p)}{\sum_{p' \in \Omega} \hat{H}(p')}
\end{equation}
The heatmap loss is simply the MSE between $H^*$ and $\hat{H}$ with mean reduction.
\begin{equation}
    L_{\text{heatmap}} = \frac{1}{|\Omega|} \sum_{p \in \Omega} ||\hat{H}(p) - H^*(p)||_2^2
\end{equation}

\subsubsection{Spatial Expectation Loss}
\label{key_dynam:appendix:spatial_expectation}
Given a descriptor $d^*$ together with a descriptor image $\mathcal{I}_D$ we can compute the 2D spatial expectation as
\begin{equation}
    \label{key_dynam:appendix:eq:spatial_expectation_pixel}
    J_\text{pixel}(d^*, I_D, \eta) = \sum_{p \in \Omega}p \cdot \tilde{H}(p; d^*, I_D, \eta)
\end{equation}
If we also have a depth image $\mathbf{Z}$ then we can define the spatial expectation over the depth channel as
\begin{equation}
    \label{key_dynam:appendix:eq:spatial_expectation_z}
    J_z(d^*, I_D, \mathcal{Z}, \eta) = \sum_{p \in \Omega}\mathbf{Z}(p) \cdot \tilde{H}(p; d^*, I_D, \eta)
\end{equation}

The spatial expectation loss is simply the L1 loss between the ground truth and estimated correspondence using
\begin{equation}
    L_{\text{spatial pixel}} = ||p^* - J_\text{pixel}(d^*)||_1
\end{equation}
We can also use our 3D spatial expectation $J_z$ to compute a 3D spatial expectation loss. In particular given a depth image $\mathbf{Z}$ let the depth value corresponding to pixel $p$ be denoted by $\mathbf{Z}(p)$. The spatial expectation loss is simply
\begin{equation}
    L_{\text{spatial z}} = ||\mathcal{Z}(p^*) - J_z(d^*, I_D, \mathcal{Z}, \eta)||_1
\end{equation}
being careful to only take the expectation over pixels with valid depth values $\mathcal{Z}(p)$.

\subsubsection{Total Loss}
The total loss is a combination of the heatmap loss and the spatial loss
\begin{equation}
    L = w_{\text{heatmap}} L_{\text{heatmap}} + w_{\text{spatial}}(L_{\text{spatial pixel}}  + L_{\text{spatial z}})
\end{equation}
where the $w$ are weights.

\subsection{Correspondence Function}
\label{key_dynam:appendix:subsec:correspondence_function}

Given a reference descriptor $d_i$ the correspondence function $g_c(\mathcal{I}_D, d^*)$ in Section \ref{key_dynam:subsec:learning_visual_representation} is computed using the spatial expectations $J_\text{pixel}(d^*, \mathcal{I}_D, \eta), J_z(d^*, \mathcal{I}_D, \mathcal{Z}, \eta)$ in Equations \eqref{key_dynam:appendix:eq:spatial_expectation_pixel}, \eqref{key_dynam:appendix:eq:spatial_expectation_z}. These spatial expectations localize the descriptor $d_i$ in either pixel space or 3D space. If in 3D we additionally use the known camera extrinsics to express the localized point in world frame.

If we define $\hat{p} = J_\text{pixel}(d^*, \mathcal{I}_D, \eta)$ as the estimated pixel location of the correspondence for descriptor $d^*$, then our visual model can additionally provide a confidence scores that $\hat{p}$ is a valid correspondence for this descriptor. The confidence is defined as the value of the unnormalized heatmap $\hat{H}$ (see Equation \eqref{key_dynam:appendix:eq:heatmap}) at the pixel location $\hat{p}$. Specifically the confidence is given by
\begin{equation}
    \hat{H}(\hat{p}; d^*, \mathcal{I}_D, \eta)
\end{equation}
Note that this value always lies in the range $[0,1]$.

\section{Training Details}
\label{key_dynam:appendix:sec:experiment_details}
This section provides details on the simulation and hardware experiments.

\subsection{Trajectory Data Augmentation}
\label{key_dynam:subsec:trajectory_data_augmentation}

Many physical systems exhibit invariances in their dynamics. For example, in the environments we consider the dynamics are invariant to translation in the $xy$ plane and rotation about the $z$ axis. In other words, if we translate or rotate our frame of reference the dynamics don't change. Encoding this invariance into our dynamics model has the potential to greatly simplify the learning problem. \cite{hogan2018data} achieves this by parameterizing the dynamics relative to the object frame, however this approach requires having access to the ground truth object frame and assumes you are dealing with a single rigid object. Another approach, taken by \citep{zeng2018learning, zeng2019tossingbot, suh2020surprising}, is to rotate the observation into a frame defined by the action. While this can work well in the setting of simple manipulation primitives (e.g. push along a straight line for $5cm$) it doesn't naturally extend to the realtime feedback setting where you are commanding actions continuously at $5-10$ Hz without returning the robot to a reference position. Since in our approach the latent-state $\bm{z}$ is a physically grounded 3D quantity we are able to encode some of this invariance by using an alternative approach based on data augmentation. Given a latent-state action trajectory $\{(\bm{y}_t, \bm{a}_t)\}$ then transforming this trajectory using a homogeneous transform $T$, which consists of $xy$ translation and $z$ rotation, yields another valid trajectory $\{(\tilde{\bm{y}}_t, \tilde{\bm{a}}_t)\} = \{(T \cdot \bm{y}_t, T \cdot \bm{a}_t)\}$. At training time we augment the training trajectories by sampling such random homogeneous transforms $T$.

\subsection{Training Details}
\label{key_dynam:appendix:subsec:training_details}

All methods used the same architecture for the dynamics model $\hat{f}_{\theta_{dyn}}$, an MLP with two hidden layers of $500$ units. All of our variants, along with the \textbf{transporter} baselines, use visual pre-training. The visual models are trained for 100 epochs and the model with the best test error is used. The dense-correspondence model uses both camera views for visual pretraining, while the transporter model uses only the images from the camera used at test time. For the dynamics learning all methods are trained for $1000$ epochs using an Adam optimizer \cite{kingma2014adam} with a learning rate of $10^{-4}$. For each method the model with the best test error was used for evaluation. Table \ref{key_dynam:table:learnable_params} details the set of learnable parameters for each method.

\begin{table}
\begin{center}
 \begin{tabular}{|c|c|} 
 \hline
 Method & Learnable Parameters $\Theta$\\
 \hline\hline
 \textbf{DS} & $\{\theta_{dyn}\}$\\
 \hline
 \textbf{SDS} & $\{\theta_{dyn}\}$ \\
 \hline
 \textbf{WDS} & $\{\theta_{dyn}, \alpha \}$ \\
 \hline
 \textbf{WSDS} & $\{\theta_{dyn}, \alpha \}$ \\
 \hline
 \textbf{Transporter} & $\{\theta_{dyn}\}$\\
 \hline
 \textbf{Autoencoder} & $\{\theta_{dyn}, \theta_\text{autoencoder} \}$\\
 \hline
\end{tabular}
\end{center}
\caption{The set of learnable parameters for the different methods during the dynamics learning phase. For our methods and transporter, these parameters don't include the weights of the visual model which remain fixed during the dynamics learning phase.}
\label{key_dynam:table:learnable_params}
\end{table}

\section{Online Model-Predictive Control}
\label{key_dynam:appendix:sec:mpc}

Following \cite{nagabandi2019deep} we use the model-predictive path integral (MPPI) approach derived in \citep{williams2015model}. Here we provide a brief overview but refer the reader to \cite{williams2015model} for more details. MPPI is a gradient-free optimizer that considers coordination between timesteps when sampling action trajectories. The algorithm proceeds by sampling $N$ trajectories, rolling them out using the learned model, computing the reward/cost for each trajectory, and then re-weighting the trajectories in order to sample a new set of trajectories. Let $H$ be look-ahead horizon of the MPC, then a single trajectory consists of state-action pairs $\{(x_t^{(k)}, a_t^{(k)} \}_{t=0}^{H-1}$. Let $R_k = \sum_{t'=t}^{t+H-1} r(x_t^{(k)}, a_t^{(k)})$ be the reward of the $k$-th trajectory. Define
\begin{equation*}
    \mu_t = \frac{\sum_{k=0}^N \left(e^{\gamma R_k} \right) a_t^{(k)} }{\sum_{k=0}^N e^{\gamma R_k}}
\end{equation*}
A filtering technique is then used to sample new trajectories from the previously computed mean $\mu_t$. Specifically
\begin{equation}
    a_t^i = n_t^i + \mu_t
\end{equation}
where the noise $n_t^i$ is sampled via
\begin{align}
    u_t^i &\sim \mathcal{N}(0, \Sigma), \hspace{0.2cm} \forall i \in {0, \ldots, N-1}, \hspace{0.2cm} \forall t \in {0, \ldots, H-1} \\
    n_t^i &= \beta u_t^i + (1-\beta) n_{t-1}^i \hspace{0.2cm} \text{ where } n_{t < 0} = 0
\end{align}
This procedure is repeated for $M$ iterations at which point the best action sequence is selected. All of our experiments we used $N = 1000, M=3, H=10, \beta=0.7$. The cost/reward in the MPC objective varied slightly between the hardware and simulation experiments, more details are provided below.

\section{Simulation Experiments}
\label{key_dynam:subsec:simulation_tasks}

To evaluate our method we consider four manipulation tasks in simulation. We use the Drake simulation environment \cite{drake} which provides both the underlying physics simulation and rendering of RGBD images at VGA resolution $640 \times 480 \times 3$. Figure \ref{key_dynam:fig:sim_images} shows image from the four simulation tasks that we consider.

\begin{itemize}[font=\bfseries]
    \item \textbf{top down camera:} This environment, depicted in Figure \ref{key_dynam:fig:sim_images} (a), consists of the sugar-box object from the YCB dataset \cite{Calli2017} laying flat on a table. The robot is represented as cylindrical pusher (shown in green) and the action $\bm{a}$ is the $x-y$ velocity of the pusher in the plane. The environment timestep is $dt = 0.1$, so the agent must command actions at 10Hz. Two cameras are placed directly above the table, facing downwards. The camera positions are offset by $90$ degrees about their z-axis. Our methods use both camera feeds for training the visual correspondence model, but only one camera feed at test time. An image from this camera is shown in Figure \ref{key_dynam:fig:sim_images} (a). All other methods use only a single camera feed at both train and test time. 
    
    \item \textbf{angled camera:} This environment is identical to \textit{top down camera} but has different camera positions. Instead of being top down the two cameras are located on adjacent sides of the table and angled at 45 degrees, see figure \ref{key_dynam:fig:sim_images} (b). The setting of angled cameras is more similar to our hardware experimental setup and is useful for comparing approaches that use pixel space vs. 3D representations.
    
    \item \textbf{occlusions:} This environment uses the same setup of task \textit{angled camera} the only difference being that the object is now laying on its side, see Figure \ref{key_dynam:fig:sim_images} (c). This, together with the angled camera position, means that occlusions become a significant factor. In particular as the box rotates through the full $360$ degrees in yaw, the sides of the box become alternately occluded or visible. The top face of the box is the only one that remains unoccluded for all poses of the object.
    
    \item \textbf{mugs (category):} This environment has the same top-down camera placements and cylindrical pusher as task \textit{angled camera}. Instead of a single object however, we use a collection of 10 different mug models and vary the color and texture on each episode. This environment tests category-level vision and dynamics generalization. Two mug instances are shown in Figures \ref{key_dynam:fig:sim_images} (d) and (e).
    
\end{itemize}

\subsection{Data Collection}
For each environment we collect a static dataset that is then used to learn the visual dynamics model. All methods have access to exactly the same dataset and the visual pretraining for our method and the \textbf{transporter} baseline is done using this same dataset. For each task the dataset is generated by collecting 500 trajectories of length 40 using a scripted random policy. The simulator timestep is $0.1$ seconds so a trajectory of length $40$ equates to $4$ seconds.

\subsection{Evaluating closed-Loop MPC performance}
For each environment we evaluate the different methods by planning to a desired goal-state image and computing the pose error (both translation and rotation) using the ground truth simulator state. Goal states are generated sampling a random control input and applying it to the environment for $15$ time steps. We further require that goal states are sufficiently far from initial states (in both translation and rotation). This generates a diverse set of initial and goal state pairs for evaluation. The simulator state is then reset to the initial state and we use closed-loop MPC to control the system to the goal state. The MPC cost function is simply the L2 distance between the latent state and the goal state. Ground truth state information is used to compute the error between the final and goal poses for the object. 

\subsection{Baselines}
\label{key_dynam:appendix:subsec:baselines}

To demonstrate the benefits of our approach over prior methods we compare against several baselines.
\begin{itemize}[font=\bfseries]
    \item \textbf{Ground Truth 3D points (GT\_3D):} This baseline is used for tasks \textit{top down camera, angled camera, occlusions} since those environment use just a single object. $\bm{z}_\text{object}$ contains ground truth world-frame 3D locations of 4 points on the object. Knowing the location of 4 points is equivalent to knowing the object pose for a rigid object. We believe that this is a strong baseline that provides an upper bound on what is achievable with our descriptor-based methods that attempt to track points on the object.
    
    \item \textbf{Transporter:} We use the \emph{Transporter} autoencoder formulation from \cite{kulkarni2019unsupervised} to pre-train a visual model. Following the original paper we use $6$ keypoints and freeze the visual model while training the dynamics model. We investigate two variants using the transporter approach. In \textbf{transporter 2D} $\bm{z}_\text{object}$ are the pixel-space locations of the keypoints. In \textbf{transporter 3D} $\bm{z}_\text{object}$ are the 3D world frame locations of the keypoints, computed from the pixel space by using the depth image together with the camera intrinsics and extrinsics. 
    
    \item \textbf{Autoencoder:} This method jointly learns the visual model and the dynamics model. Specifically we jointly train a convolutional autoencoder together with a forward dynamics model. The loss is a combination of the dynamics loss, Equation \eqref{key_dynam:eq:dynamics_cost_function}, and an image reconstruction loss, which penalizes the L2 distance between the reconstructed and actual images. Note that this is exactly the autoencoder baseline from \cite{yan2020learning}. Following \cite{yan2020learning} images are downsampled to $64 \times 64$ before being passed into the network. The encoder architecture contains 6 2D convolutions with kernel sizes $[3,4,3,4,4,4]$, strides $[1,2,1,2,2,2]$ and filter sizes $[64, 64, 64, 128, 256, 256]$. Leaky ReLU activations are added between the convolutional layers. The final output is flattened and passed through a fully-connected layer to form the latent-state $\bm{z}_\text{object}$. We experimented with different dimensions for $\bm{z}$ from $16$ to $64$ and found that $64$ worked best. Hence a $64$ dimensional latent state is used for all experiments. The decoder follows the one in \cite{hafner2018learning} and consists of a dense layer followed by 4 transposed convolutions with kernels size 4 and stride 2 which upscales the output image to $64 \times 64$.
\end{itemize}

\section{Hardware Experiments}
\label{key_dynam:appendix:hardware}

\subsection{Hardware Setup}
\begin{figure}
\centering
\includegraphics[width=\textwidth]{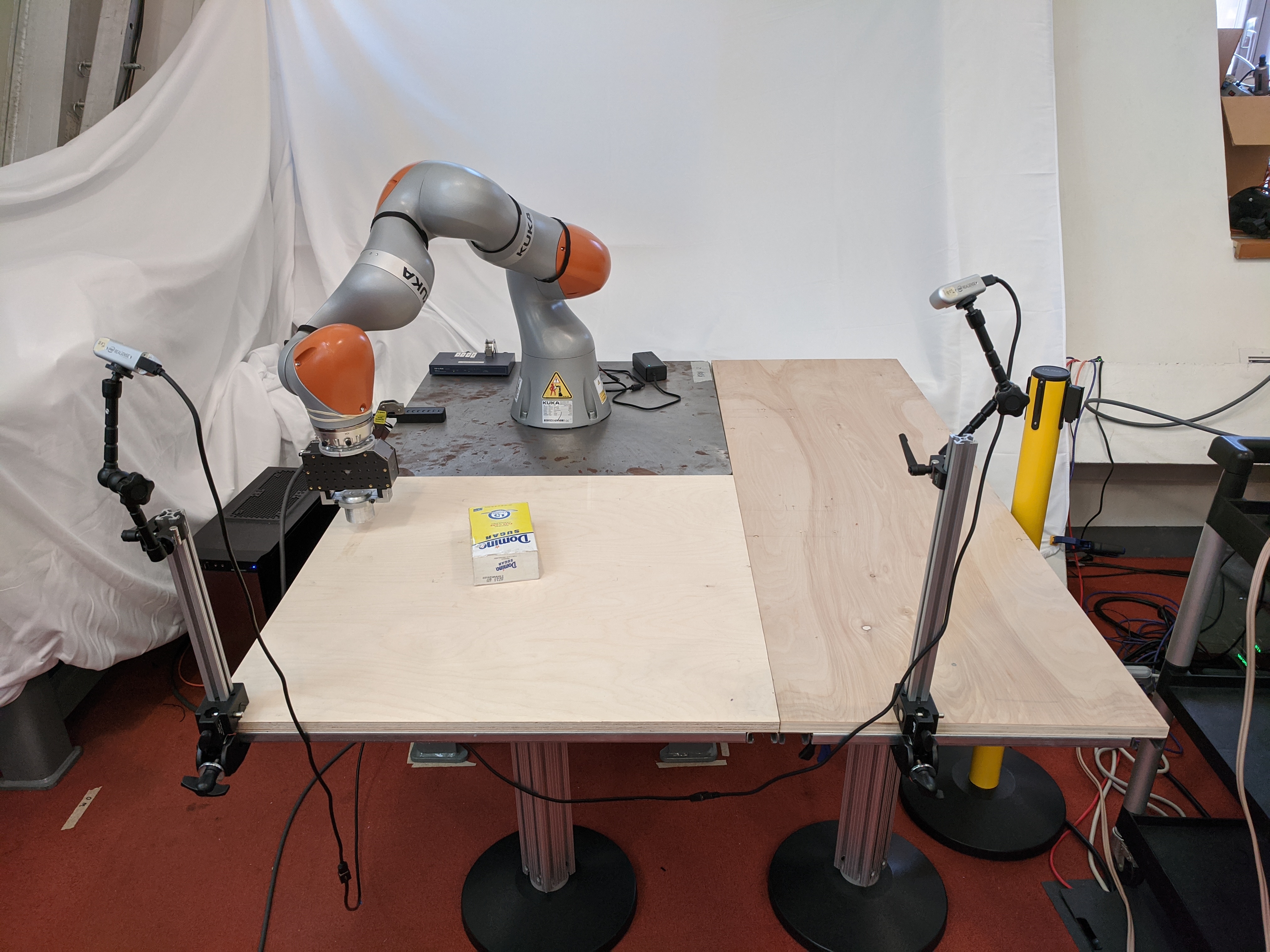}
\caption{\label{key_dynam:fig:hardware_overview} Overview of our experimental setup, including the two external Realsense D415 cameras. Images from both cameras are used to train the dense-descriptor model, while only the right camera is used at runtime to localize the keypoints on the object.}
\end{figure}

We used a Kuka IIWA LBR robot with a custom cylindrical pusher attached to the end-effector to perform our hardware experiments, see Figure \ref{key_dynam:fig:hardware_overview}. RGBD sensing was provided by two RealSense D415 cameras rigidly mounted offboard the robot and calibrated to the robot's coordinate frame. To enable effective correspondence learning between views, it is ideal to have views with \textit{some} overlap such that correspondences exist, but still maintain different-enough viewpoints from each camera. At test time only a single camera is used to localize the dense-descriptor keypoints. The robot is controlled by commanding end-effector velocity in the $xy$ plane at 5Hz. A high-rate Jacobian space controller consumes these 5Hz end-effector velocity commands and closes the loop to command the robot's joint positions at 200Hz.

\subsection{One-Shot Imitation Learning}
\begin{figure}[h]
\centering
\includegraphics[width=\textwidth]{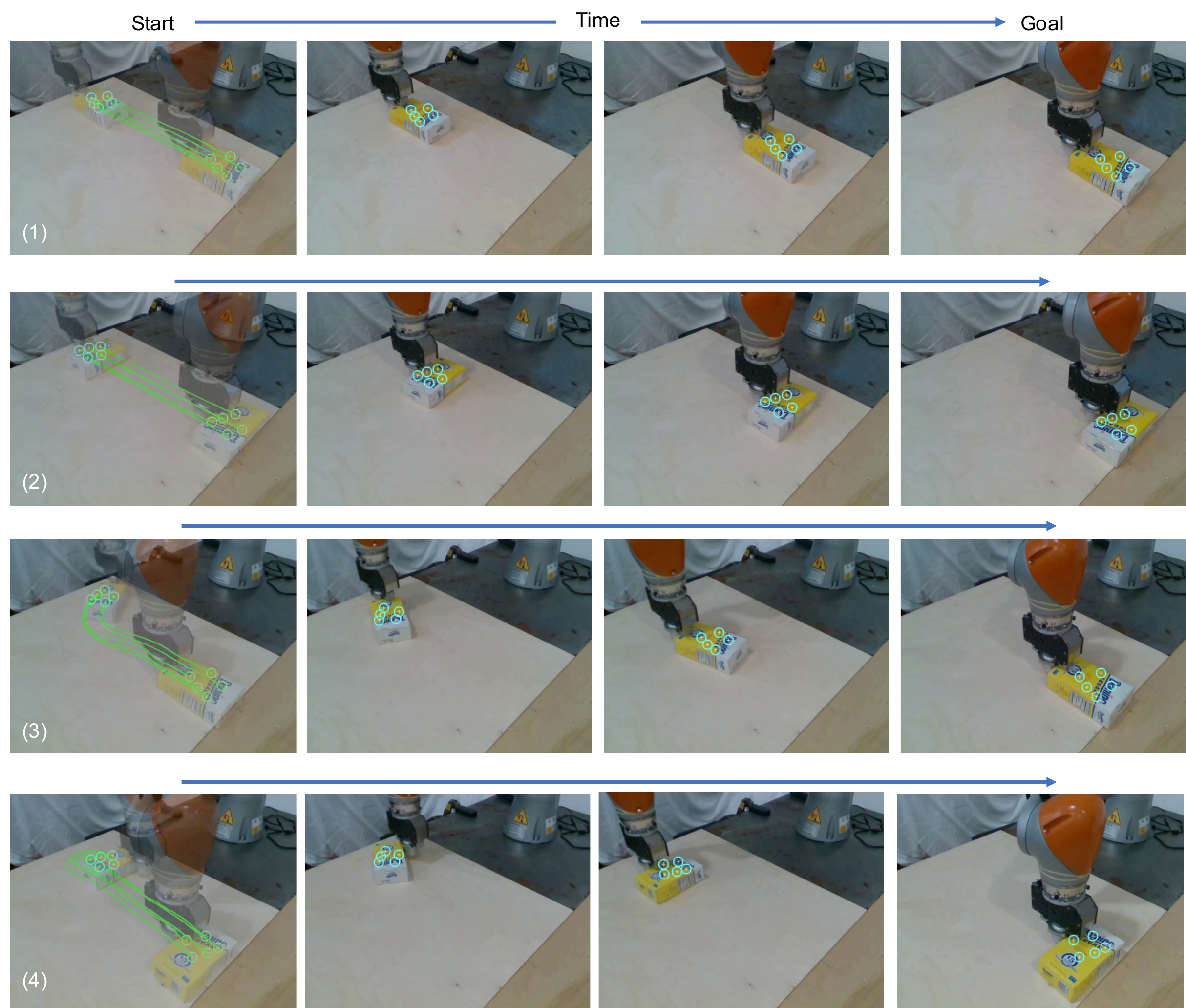}
\caption{\label{key_dynam:fig:hardware_demonstration_trajectories} The 4 demonstration trajectories used for the hardware experiments. The left image of each row shows the starting position blended with the goal position. The \textbf{SDS} keypoints are shown in teal for each frame. The green lines show the paths followed by the keypoints moving from the starting position to the final position. The right image of each row shows the final/goal position.}
\end{figure}

Although our learned dynamics model together with online MPC is able to plan over short to medium horizons, we can track much longer horizon plans by providing a single demonstration and using a trajectory tracking cost in our MPC formulation. This demonstration can in principle come from any source, in our case we used a human teleoperating the robot. We capture observations throughout the trajectory at $5Hz$ resulting in a trajectory of observations $\{\bm{o}_t^*\}_{t=0}^T$. Using our visual model we convert these observations into keypoints $\{\bm{y}_t^*\}_{t=0}^T$ and latent state $\{\bm{z}_t^*\}_{t=0}^T$ trajectories. These trajectories are used to guide the MPC. In particular the cost/reward function in the MPC is 
\begin{equation}
    r(z_t, a_t) = - ||z_t - z_t^*||_2^2
\end{equation}
This trajectory cost allows us to accurately track long horizon plans (where the demonstrations are as long as 15 seconds) using an MPC horizon of $2$ seconds. The four demonstrations trajectories used in the hardware experiments are illustrated in Figure \ref{key_dynam:fig:hardware_demonstration_trajectories}.

\subsection{Results}

\begin{figure}[h!]
\centering
\includegraphics[width=\textwidth]{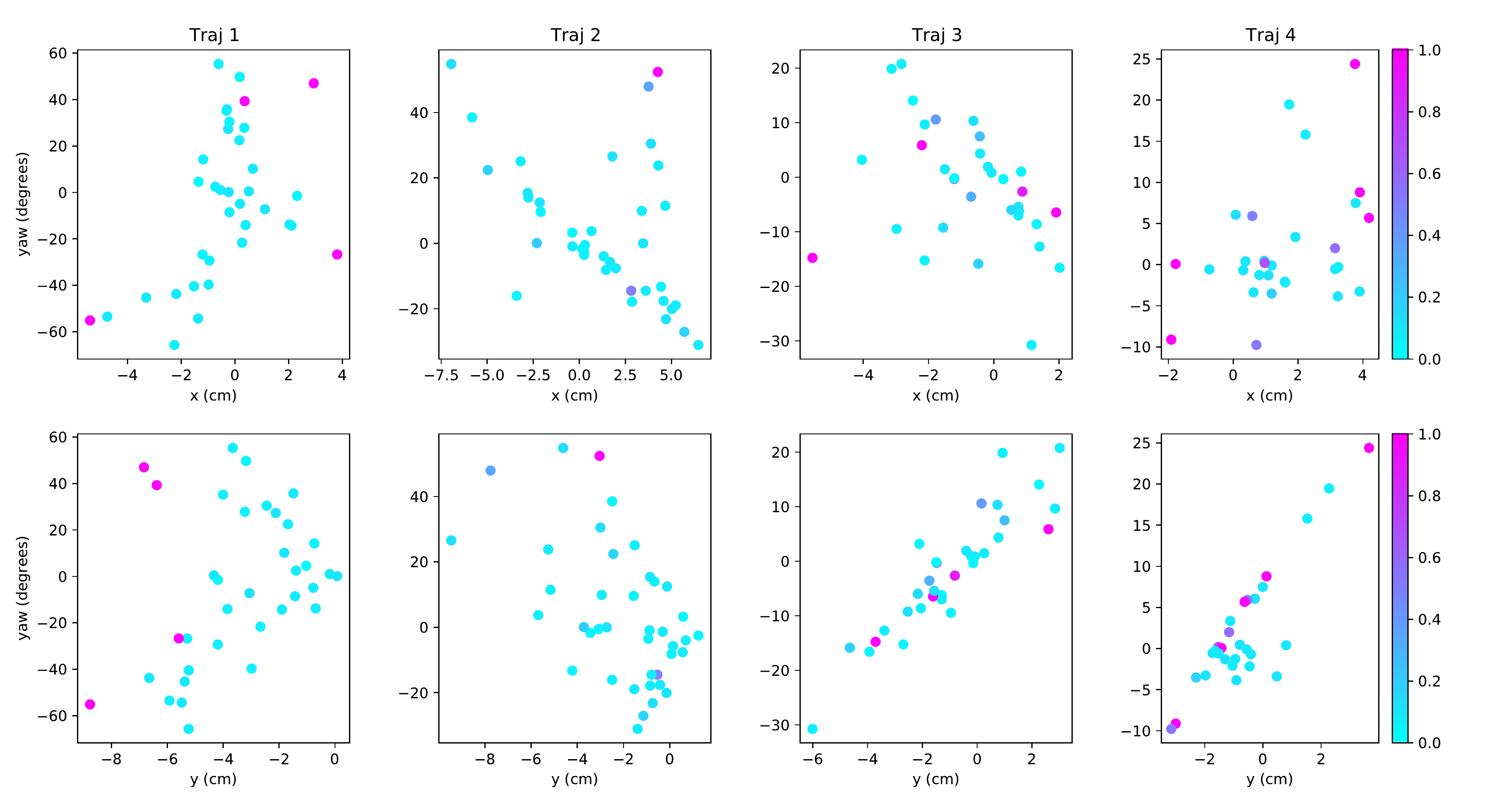}
\caption{\label{key_dynam:fig:hardware_quant_results_both_axes} Scatter plots show results of our approach on the four different reference trajectory tracking tasks. The axes of the plots show the deviation of the object starting pose from the initial pose of the demonstration. The top row shows the deviation in the $x$ and $yaw$ axes, while the bottom shows the deviation in the $y$ and $yaw$ axes. Axes are illustrated in Figure \ref{key_dynam:fig:hardware_results}. The color indicates the distance between final and goal poses, lower cost is better. The numerical value is computed as $\text{cost} = \frac{\Delta \text{pos}}{3} + \frac{\Delta \text{angle}}{30}$ where $\Delta \text{pos}, \Delta \text{angle}$ are the translational (in centimeters) and angular (in degrees) errors between the object's final position and the goal position. The costs are rescaled and plotted in the range $[0,1]$. The various reference trajectories, shown in Figure \ref{key_dynam:fig:hardware_demonstration_trajectories}, are of different difficulties, as reflected by the different regions of attraction of the MPC controller.  Videos of the closed-loop rollouts can be found at \href{https://sites.google.com/view/keypointsintothefuture}{project page}.}
\end{figure}

In this section we provide more details on the hardware experiments from Section \ref{key_dynam:subsec:hardware}. Figure \ref{key_dynam:fig:hardware_quant_results_both_axes} expands on Figure \ref{key_dynam:fig:hardware_results} showing the region of attraction of our MPC controller when attempting to stabilize the four different trajectories shown in Figure \ref{key_dynam:fig:hardware_demonstration_trajectories}. We define a trajectory as a success if the final object position is within $3$ cm and $30$ degrees of the goal position. Given that during trials we explicitly chose initial conditions to test the region of attraction of the MPC controller, success rates are not particularly meaningful, as the success rate depends on the initial condition. Table \ref{key_dynam:table:hardware_quant_results} shows the average translational and angular errors among successful trajectories.

\begin{table}[h!]
\begin{tabularx}{\textwidth}{XRRRR}
\toprule
Trajectory & pos (cm)  & angle $^\circ$ & success rate & num trials \\
\midrule
1 & $1.23 \pm 0.55$ & $5.25 \pm 3.99 $ & $87.5\%$ & $40$ \\
2 & $1.10 \pm 0.319$ & $7.32 \pm 4.08$ & $89\%$ & $36$\\
3 & $1.16 \pm 0.58$ & $3.80 \pm 2.54$ & $73\%$ & $33$\\
4 & $1.44 \pm 0.30$ &  $9.73 \pm 5.48$ & $65\%$ & $29$ \\
\bottomrule
\end{tabularx}
\vspace{0.2cm}
\caption{Quantitative results of hardware experiments. A trial is considered a success rate if the final object position was within $3$ cm and $30$ degrees of the goal pose. Note that, as shown in Figure \ref{key_dynam:fig:hardware_quant_results_both_axes} the initial conditions were intentionally chosen to test the region of attraction of our controller, thus the success rates are not meaningful in and of themselves and are included only for completeness. The pos (cm) and angle columns show the deviation of the final object position from the target. Note that the mean and standard deviation are only calculated over the successful trials. This serves to give a sense of the accuracy that can be achieved by using our closed-loop MPC controller.}
\label{key_dynam:table:hardware_quant_results}
\end{table}

\end{document}